\documentclass[11pt]{article}

\usepackage[]{acl}

\usepackage{times}
\usepackage{latexsym}
\usepackage[T1]{fontenc}
\usepackage[utf8]{inputenc}
\usepackage{microtype}

\usepackage{graphicx}
\usepackage{amsmath}
\usepackage{booktabs}
\usepackage{multirow}
\usepackage[normalem]{ulem}
\usepackage{enumitem}
\usepackage{amsmath}
\usepackage{amsfonts}

\title{GPTFace: Generative Pre-training of Facial-Linguistic Transformer by Span Masking and Weakly Correlated Text-image Data}

\author{Yudong Li\footnotemark[2], 
        Hao Li\footnotemark[2], 
        Xianxu Hou, 
        Linlin Shen\footnotemark[1] \\
   Shenzhen University \\
   }

\begin{document}
\maketitle

\begin{abstract}
Compared to the prosperity of pre-training models in natural image understanding, the research on large-scale pre-training models for facial knowledge learning is still limited. Current approaches mainly rely on manually assembled and annotated face datasets for training, but labeling such datasets is labor-intensive and the trained models have limited scalability beyond the training data. To address these limitations, we present a generative pre-training model for facial knowledge learning that leverages large-scale web-built data for training. We use texts and images containing human faces crawled from the internet and conduct pre-training on self-supervised tasks, including masked image/language modeling (MILM) and image-text matching (ITM). During the generation stage, we further utilize the image-text matching loss to pull the generation distribution towards the control signal for controllable image/text generation. Experimental results demonstrate that our model achieves comparable performance to state-of-the-art pre-training models for various facial downstream tasks, such as attribution classification and expression recognition. Furthermore, our approach is also applicable to a wide range of face editing tasks, including face attribute editing, expression manipulation, mask removal, and photo inpainting.
\end{abstract}

\section{Introduction}

Over the past few years, there has been extensive research on human face analysis and generation using deep learning-based methods. However, these methods typically require supervised data to establish a link between the model and human perception. In previous research, state-of-the-art methods \cite{yang2020hierarchical, chen2021understanding, hou2022textface} have relied on datasets that are annotated with knowledge related to human faces. For example, CelebA \cite{liu2015deep} annotates 40 facial attributes, FairFace \cite{karkkainen2021fairface} annotates race, age, and gender, and Multi-Modal CelebA-HQ \cite{xia2021tedigan} provides textual descriptions for face images. However, constructing such datasets is challenging due to the need for large amounts of manually labeled data. Moreover, these datasets are often annotated separately for different aspects and in various formats, making it difficult to combine them effectively.

Recent progress in machine learning has led to impressive success in vision models trained with natural language supervision signals \cite{radford2021learning, cho2021unifying}. With a large amount of text-image data available on the Internet, visual concept representations learned directly from text provide broader sources of supervision and achieve better zero- or few-shot learning performance compared to fixed predetermined object categories. Several studies have explored the potential of vision-language pre-training based on publicly available datasets \cite{schuhmann2021laion, changpinyo2021conceptual, gu2022wukong}.

\begin{figure}[t]
\centerline{\includegraphics[width=\columnwidth]{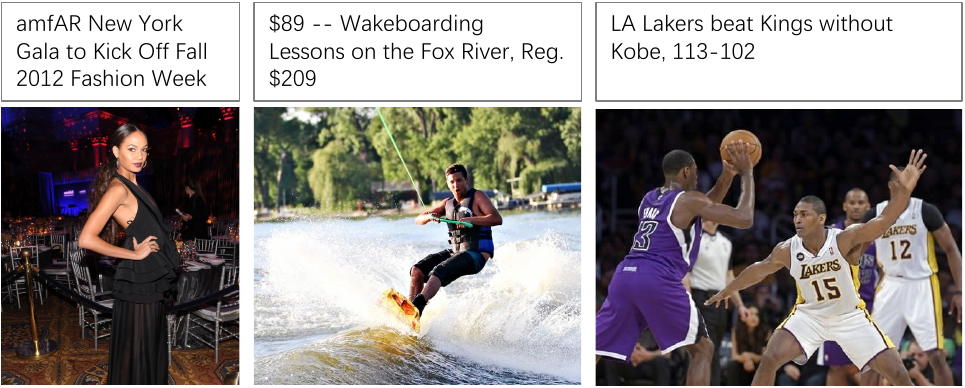}}
\caption{The LAION-FACE dataset \cite{zheng2022general}, which is a subset of LAION-400M \cite{schuhmann2021laion}, contains samples of facial image-text pairs that showcase the challenge of utilizing text information for human faces obtained from internet text-image pairs as the related text information is often uninformative.}
\label{dataset}
\end{figure}

In the field of facial analysis, there are also pre-training models \cite{zheng2022general, li2022masked} that learn from large-scale face-related images and texts. However, these models are primarily designed for text-image retrieval, which can be applied to facial classification or parsing, but are challenging to adapt for generative tasks. The primary obstacle is the weak semantic association between the Internet text-image data from the perspective of the face domain. As illustrated in Figure~\ref{dataset}, it is difficult to extract useful facial information from the given text. This lack of strong correlation between facial text-image pairs makes it challenging for generative pre-training models to learn effectively.

To address the aforementioned challenges, we propose a generative pre-training model, called GPTFace, to learn facial knowledge from weakly correlated text-image data in the face domain. GPTFace can simultaneously perform both face analysis and generation tasks. Specifically, we train a generative model using a shared model structure and parameters from both masked image/language modeling (MILM) and image-text matching (ITM) tasks. Previous works such as BERT \cite{devlin2018bert} and MAE \cite{he2022masked}, typically use MILM tasks for representation learning, which are modeled from masked text and images and then transferred to downstream tasks. In this paper, inspired by SpanBERT \cite{joshi2020spanbert}, we propose span image-text masking that is effective for generative pre-training. Additionally, for controllable generation, we propose a simple but effective method that uses ITM supervision to update the encoder parameters, allowing for the manipulation of the output image/text distribution based on the input text/image.

Compared to existing visual-linguistic generative methods, which typically use texts or images as direct supervision signals, our approach jointly models texts and face images with shared parameters and utilizes the ITM loss to guide generation. We have found that this strategy is effective in learning from weakly correlated data. In addition, GPTFace focuses on the face domain, leading to faster convergence than existing general domain pre-training methods. Our experimental results demonstrate that GPTFace is suitable for various face-related scenarios, such as expression and attribute editing, occlusion removal, and face inpainting. Furthermore, for facial analysis tasks such as facial attribute classification and expression recognition, our model achieves competitive performance comparable to state-of-the-art large-scale pre-training models. In summary, our contributions can be outlined as follows:

\begin{itemize}[itemsep=2pt,topsep=2pt,parsep=2pt]
\item We introduce GPTFace, the first facial-linguistic generative pre-training model that learns general face knowledge from large-scale weakly correlated text-image data.

\item We propose a novel gradient-based method using image-text matching guidance to achieve controllable generation.

\item The experimental results demonstrate that our model can perform various face-related tasks and achieves outstanding performance.
\end{itemize}

\section{Related Work}

\textbf{Text-guided Face Editing} aims to manipulate specific attributes of a face image based on text descriptions while keeping other attributes unchanged. Previous methods typically manipulate the latent space of pre-trained GANs to achieve editing in the image space. To align the representations of language with GAN latent spaces, recent methods \cite{hou2022textface, xia2021tedigan, wang2021faces, avrahami2022blended} train the text embedding network using human-annotated face descriptions \cite{jiang2021talk, xia2021tedigan, sun2021multi}.
More recently, several approaches \cite{xia2021towards, patashnik2021styleclip, sun2022anyface} have used the contrastive language-image pre-training model (CLIP) \cite{radford2021learning} as a text encoder to achieve face manipulation with pure text descriptions. In these approaches, the image and text modules are trained separately, and then extra efforts are required to align the image and text representations, which restricts the models' generalization capability. In contrast, our approach jointly encodes texts and images in a shared discrete space, enabling both image and text editing in a unified framework. This allows for greater flexibility and enhances the model's generalization capability.

\textbf{Vision-language Generative Pre-training.} Transformer and its variants \cite{vaswani2017attention, child2019generating, lee2021fnet} have been used as powerful backbone networks for state-of-the-art language models \cite{devlin2018bert, radford2019language, liu2019roberta}. Drawing inspiration from the success of language models, transformer and the pretraining-finetuning paradigm have also been widely adopted for vision and cross-modal tasks \cite{kim2021vilt, bao2021beit, cho2021unifying, radford2021learning}. To utilize transformers for image modeling, current approaches typically represent images as sequences and generate images utilizing the same autoregressive decoding process as text generation.

For example, DALL-E \cite{ramesh2021zero} formulates the text-to-image synthesis as a sequence-to-sequence task, where image tokens are learned through discrete VAE \cite{van2017neural}. ERNIE-ViLG \cite{zhang2021ernie} achieves bidirectional text-and-image generation with sequence modeling and adopts pre-trained VQGAN \cite{esser2021taming} image tokenization. However, the assumption that images and texts are strongly correlated is invalid in the face domain of large-scale text-image datasets \cite{zheng2022general}. These pre-training methods trained using large-scale natural images do not perform well in face generation. To mitigate this problem, Talk2Face \cite{li2022talk2face} converts supervised data labels into text for training. However, constructing text data directly from labels results in limited textual diversity. In this work, we model images and text independently and learn their relationship through the image-text matching task, utilizing weakly correlated text-image data.

\section{Approach}

\begin{figure*}[h]
\centerline{\includegraphics[width=\textwidth]{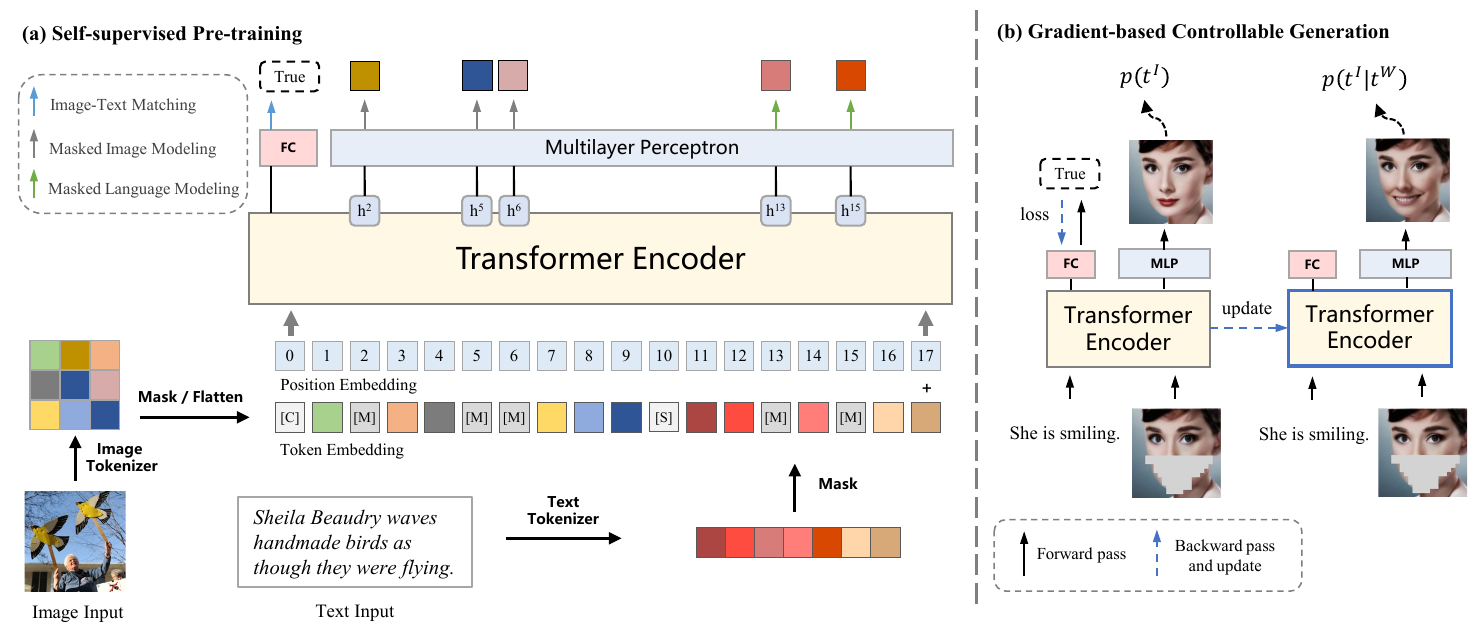}}
\caption{Overview of our approach. During pre-training, our model learns to generate text and images using shared parameters and masked image/language modeling tasks. The image-text matching objective helps to model the association between text and image, and guide the generator to achieve conditional image/text generation during inference.}
\label{overview}
\end{figure*}

In this section, we introduce the model architecture, input/output format, pre-training tasks and the generation process of GPTFace. Figure~\ref{overview} illustrates an overview of our approach.

\subsection{Model Architecture}

\textbf{Input Representations.} 
We represent text and image as discrete sequences in the same format. For images, we use the encoder of pre-trained VQGAN \cite{esser2021taming} to map and quantize input image $x \in \mathbb{R}^{C\times W\times H }$ into discrete image tokens $t^I = [t^I_1,\dots,t_n^I] \in  \boldsymbol{C}$, where $n$ is the number of image patches and $\boldsymbol{C}$ is the codebook. For text, we use WordPiece \cite{wu2016google} to tokenize text into uncased word tokens $t^W = [t^W_1,\dots,t_m^W] \in \boldsymbol{V}$, where $m$ is the length of text and $\boldsymbol{V}$ is the vocabulary.

Given an image-text pair, the discrete representations are obtained as above. 
We then concatenate them with special tokens [CLS] and [SEP], $t = [t_{CLS},t^I_1,\dots,t_n^I,t_{SEP},t^W_1,\dots,t_m^W] \in \boldsymbol{C \cup V}$. The start token [CLS] is placed at the beginning of the sequence and the separate token [SEP] marks the boundary between text and image tokens. The sequence is then linearly projected to obtain the token embedding $E_t = [E_{CLS},\dots,E^W_{m}]$. We adopt the standard learnable 1D position embedding $E_{pos}$, and add it to the token embedding.

\textbf{Backbone Network.}
To process both image and text data simultaneously, we employ a shared transformer encoder \cite{vaswani2017attention}. Following recent state-of-the-art transformer implementations \cite{xue2021mt5, du2021glam, thoppilan2022lamda}, we move Layer Normalization (LN) layers to the input of each sub-block \cite{radford2019language} and adopt the Gated Linear Unit (GLU) \cite{dauphin2017language} as the feed-forward network. The input of the first transformer block is $H^0 = E_t + E_{pos}$, the output of \textit{l}-th transformer block is computed via the following equations:
\begin{equation}
    \hat{H^l} = Attention(LN(H^{l-1})) + H^{l-1}
\end{equation}
\begin{equation}
    H^l = GLU(LN(\hat{H^l})) + \hat{H^l}
\end{equation}
the output $H^L$ of the last transformer block contains encoded representations for each token,
\begin{equation}
    H^L = [h_{CLS},h^I_1,\dots,h_n^I,h_{SEP},h^W_1,\dots,h_m^W].
\end{equation}
where $L$ represents the number of transformer blocks.

\subsection{Self-supervised Pre-training}

To learn from weakly correlated text-face data, our model is jointly optimized by three self-supervised tasks: masked image modeling on images, masked language modeling on text, and image-text matching on image-text pairs.

\textbf{Masked Image Modeling} has been widely used in recent pre-training models for visual representation learning, e.g., BEiT \cite{bao2021beit}, MAE \cite{he2022masked} and iBOT \cite{zhou2021image}. Current approaches adopt a strategy of randomly masking a certain percentage of image patches, which has proven to be an effective pre-training approach for classification tasks. However, the random masking strategy only requires the model to predict masked tokens based on their immediate neighbors, which may not be effective in restoring image masks that span a large number of blocks.

To use masked image modeling for generative pre-training, we propose an image span masking strategy. The image span masking strategy begins by selecting a random seed token and then iteratively masks tokens around the seed token until the pre-set budget number is reached. Given that image span masking is more challenging for image reconstruction, we set the masking budget to a small value (15\%) initially and gradually increase it as training progresses (1\% every 20,000 steps), with an upper limit of 65\%. The masked image tokens are replaced by a special token [MASK].

\begin{figure}[t]
\centerline{\includegraphics[width=\columnwidth]{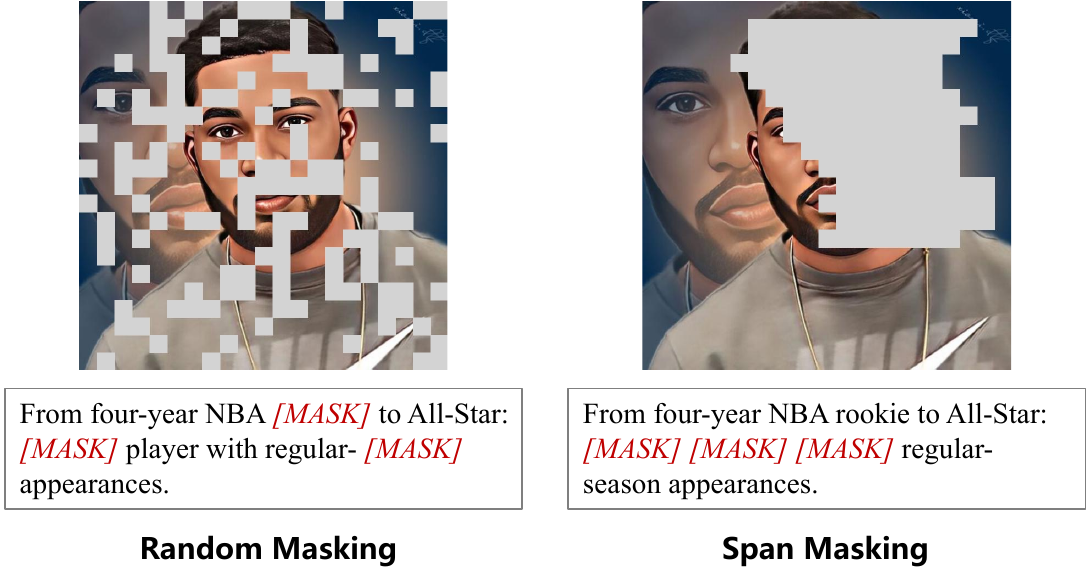}}
\caption{Visual examples of random masking and span masking on a text-image pair.}
\label{masking}
\end{figure}

\textbf{Masked Language Modeling.}
We follow the random masking strategy of BERT \cite{devlin2018bert} by randomly masking 20\% of the text tokens. To enhance generative performance, we do not employ data augmentation techniques such as random replacement and deletion. Additionally, we also adopt the span masking technique for text tokens, which involves masking a contiguous sequence \cite{joshi2020spanbert}. Figure~\ref{masking} provides a visual comparison of random masking and span masking on images and text.

Masked text and image tokens are replaced with the special token [MASK]. As our input sequence includes both image and text samples, we employ a shared softmax layer to predict the image/text tokens for each mask position $\mathcal{M}$ based on the transformer output $H^L$. Therefore, the above two tasks are able to use a shared training objective, i.e., masked image/language modeling (MILM), to optimize the following likelihood:

\begin{equation}
    \mathcal{L}_{MILM} = -\sum_{m\in \mathcal{M} } \textrm{log}\ p(t_m| t_{\setminus  \mathcal{M}})
\end{equation}

\noindent where $\mathcal{M}$ denotes masked image/text positions, and $t_{\setminus  \mathcal{M}}$ represents the remaining tokens that are not masked.

\textbf{Image Text Matching.}
In order to capture the relationship between images and text, we pre-train the model using a binarized image-text matching (ITM) task. Following ViLT \cite{kim2021vilt}, we randomly select text and image pairs for each training example, with 50\% of the pairs being aligned and the other 50\% of the text being replaced with a different sentence from the corpus.

We use the representation corresponding to the [CLS] token $h^L_{CLS}$ as the aggregate sequence representation. We then formulate image-text matching as a binary classification task and employ a single fully connected layer as ITM head to project the output feature to binary class logits $q$. We compute negative log-likelihood loss as:
\begin{equation}
    \mathcal{L}_{ITM} = -\textrm{log}\ p(q| t^I, t^W)
\end{equation}

\noindent where $t^I$ and $t^W$ denote the image and text tokens, respectively. $q=1$ if the input text-image pair is matched.

\begin{table*}[ht!]
\centering
\small
\caption{Comparison of settings and applications with existing pre-training transformer models. In contrast, GPTFace is trained more efficiently and can be used for a variety of applications. CLS: classification. CLR: contrastive learning. LM: language modeling. }
\begin{tabular}{l|lllll}
\toprule
\multicolumn{1}{c|}{\multirow{2}{*}{Model}} & \multicolumn{5}{c}{Pre-training Settings} \\
\multicolumn{1}{c|}{} & Dataset & Data Size & Epochs & Tasks & Devices \\ \midrule
ViT \cite{dosovitskiy2020image} & ImageNet-22K & 14M & 14 & CLS & 230 TPU-v3 days  \\
CLIP \cite{radford2021learning} & WIT & 400M & 32 & CLR & 256 V100 GPUs \\
DALL-E \cite{ramesh2021zero} & Hybrid & 250M & 600 & LM & 64 V100 GPUs  \\
MAE \cite{he2022masked}& ImageNet-22K & 14M & 800 & MIM & 128 TPU-v3 cores   \\
BEiT \cite{bao2021beit}& ImageNet-22K & 14M & 800 & MIM & 16 V100 GPUs   \\
FaRL \cite{zheng2022general}& LAION-FACE & 20M & 16 & MIM, CLR & 32 V100 GPUs  \\ \midrule
GPTFace(ours) & LAION-FACE & 20M & 10 & MILM, ITM & 8 A100 GPUs  \\ \bottomrule
\end{tabular}
\label{pretrain_cmp}
\end{table*}

\subsection{Gradient-based Controllable Generation}

In the pre-training phase, GPTFace is trained to learn the joint distribution of texts and face images, as well as the complex relationship between the two modalities. However, as the model is trained on weakly correlated text-image data, directly sampling for span-masked regions can be challenging. To overcome this issue, we use the ITM gradient during the inference phase to tune the generation distribution. This allows for the controlled generation of text and images.

To illustrate our approach, we consider the task of text-guided image inpainting. Given a sequence of image tokens $t^{I} = \{t^I_\mathcal{M},  t^I_{\setminus \mathcal{M}} \}$, the goal is to use text tokens $t^W = [t^W_1, \dots, t^W_m]$ to control the distribution of image generation as $p(t^I| t^W)$. Our pre-trained model is capable of predicting the unconditional probability of the masked tokens as follows:

\begin{equation}
    p(t^I) = \prod_{m \in \mathcal{M}} p(t^I_m|t^I_{\setminus \mathcal{M}}, \theta)
\end{equation}

\noindent where $\theta$ is the model parameters. In addition, Since our model employs shared parameters, it is pre-trained on the ITM task to model $p(q|t^I, t^W, \theta)$.

We adopt a non-autoregressive decoding method to synthesize an image in a fixed number of steps. In order to control the output of masked image model, we first compute the ITM loss between the input image and text at every generation step to obtain the gradient. Then, we update the model parameters with step size $\lambda$ toward the direction of higher log-likelihood, indicating that the text and image are better aligned.

\begin{equation}
    \theta \leftarrow (1-\lambda)\theta + \lambda \frac{\partial \textrm{log}\ p(q=1|t^I,t^W,\theta) }{\partial \theta}
    \label{eq6}
\end{equation}

The next token is then sampled from the updated distribution, which is more likely to possess the attributes described in the text, i.e., $p(t^I| t^W) \propto p(t^I) p(q=1|t^I, t^W)$.

While our approach can guide the generator towards a specified direction during inference, gradient accumulation may lead to the generation of unrealistic examples when the generator moves into low-probability regions \cite{szegedy2014intriguing, dathathri2019plug}. To address this issue, we restore the model parameters after each sampling step. This ensures that the shifted distribution does not deviate from the regions with high $p(t^I)$. Notably, since both texts and images are represented using a shared discrete format, our approach is consistent for controllable generation of both modalities. In contrast to existing gradient-based controllable approaches \cite{nguyen2017plug, dathathri2019plug} that often require an external model to compute gradients, we leverage the ITM loss, which shares parameters with the generative model. This eliminates the need for an additional model, simplifying the generation process.

Existing transformer-based generative pre-training works, like GPT2 \cite{radford2019language}, Dall-E \cite{ramesh2021zero} and ERNIE-ViLG \cite{zhang2021ernie}, mainly adopt left-to-right sequence modeling. These models employ autoregressive decoding, where tokens are generated sequentially based on previously generated output. However, this process is time-consuming as each image or text requires the same number of inferences as the number of patches or length of a sentence.

In contrast, our model employs bi-directional modeling and non-autoregressive decoding, allowing for the generation of multiple tokens during each inference step. To enable efficient and high-quality generation, we draw inspiration from MaskGIT \cite{chang2022maskgit} and propose an iterative decoding method for our span masked pre-training tasks. Given a sequence of masked tokens $t_{\mathcal{M}}$ and its conditional input $t_{c}$, decoding $k$ tokens requires $|\mathcal{M}| / k$ iterations. More specifically, for iteration $i$, our decoding process is as follows:

\begin{itemize}[itemsep=2pt,topsep=2pt,parsep=2pt]
\item [1.] \textbf{Forward and Update.} Given the masked tokens $t^i_{\mathcal{M}}$, we compute the ITM loss and then update parameters as Eq~\ref{eq6}. 
\item [2.] \textbf{Recompute and Sample.} We recompute to obtain the predictions for each mask position with the updated encoder. At each masked position $j$, we sample a token $t_j$ based on model's prediction. The token's corresponding logit is used as a ``confidence'' score, indicating the model's belief in this prediction.
\item [3.] \textbf{Output.} After sampling tokens for each masked position, we select the top $k$ tokens with the highest confidence from the candidate token positions adjacent to the span boundary as output. These tokens are inserted into $t^{i+1}_{\mathcal{M}}$ as input for the next iteration.
\end{itemize}

The number of decoding iterations can be shortened by modifying the hyperparameter $k$. In theory, only one iteration is required when setting $k = |\mathcal{M}|$. However, it is usually difficult for the model to produce accurate results in a single inference, thus we empirically set $k = 8$.

\section{Experiments}

\subsection{Pre-training Setup}

\textbf{Dataset.}
For pre-training, we use a large-scale image-text dataset LAION-FACE \cite{zheng2022general}, which contains about 20 million image-text pairs. The dataset was curated by applying a face detector to filter face images from the LAION-400M dataset \cite{schuhmann2021laion}. In this dataset, the text descriptions that correspond to face images are frequently unrelated to faces (see Figure~\ref{dataset}). Notably, the text descriptions associated with the face images in this dataset are often unrelated to the faces themselves (see Figure~\ref{dataset}). Furthermore, since the face detector may produce false alarms, some images in the dataset do not contain faces.

\textbf{Tokenizer.}
To tokenize images, we pre-train VQGAN on LAION-FACE dataset with a codebook $\boldsymbol{C} = 8,192$ and factor $f = 16$. Each image is resized to $336 \times 336$ resolution and tokenized into 441 tokens. For text tokenizer, we follow the BERT English uncased vocabulary, which contains $\boldsymbol{V} = 30,522$ tokens. The maximal text input length is set to 64.

\textbf{Configuration.} We use a 12-layer transformer with 768 hidden sizes, and 12 attention heads. The intermediate size of feed-forward networks is 3,072. With approximately 110 million parameters, our model is comparable in size to BERT-base and ViT-base.

\textbf{Hyperparameters.} We train our model from random initialization, running it for 1,000,000 steps with a batch size of 192. AdamW \cite{loshchilov2018decoupled} with $\beta 1 = 0.9$, $\beta 2 = 0.999$ is employed for optimization. The learning rate is initialized with $2e^{-4}$ and linearly warmuped to $1e^{-3}$. The pre-training is conducted on 8 Nvidia Tesla A100 40GB GPUs.

\subsection{Comparison with Other Pre-training Models}

We compare GPTFace with recent transformer-based pre-training models using identical model frameworks and comparable parameters (around 110M). As shown in Table~\ref{pretrain_cmp}, our model achieves comparable performance with a minimal number of epochs and devices, pre-training on a single machine. This is due to our model's focus on the face domain and the use of various pre-training tasks to learn quickly from texts and images. In addition, the other models were designed independently for representation or generation. In contrast, the versatility of our model makes it suitable for a wide range of application scenarios.

To evaluate the performance of our model on downstream tasks, we adapt GPTFace to facial analysis tasks, including facial attributes classification and expression recognition. For facial attributes classification, we use CelebA \cite{liu2015deep}, which is annotated with 40 binary attribute labels and consists of 162,770 training samples and 19,867 testing samples for evaluation. For expression recognition, we adopt RAF-DB \cite{li2017reliable}, which contains a training set of 12,271 faces and a testing set of 3,068 faces for experiments. We employ the same experimental setup as FaRL for all models in the comparison. The results on CelebA are obtained from the FaRL paper, while we evaluate the results on RAF-DB.

\begin{table}[h]
\centering
\small
\caption{Accuracy comparison with state-of-the-art pre-training transformer models on facial attribute classification (CelebA) and expression recognition (RAF-DB).}
\resizebox{1.00\columnwidth}{!}{
\begin{tabular}{l|ccc|ccc}
\toprule
 & \multicolumn{3}{c|}{CelebA} & \multicolumn{3}{c}{RAF-DB} \\
\multicolumn{1}{c|}{} & 1\% & 10\% & 100\% & 20\% & 50\% & 100\% \\
\#sample & 1627 & 16277 & 162770 & 2454 & 6135 & 12271 \\ \midrule
ViT \cite{dosovitskiy2020image} & 89.20 & 90.21 & 90.99 & 76.96 & \textbf{83.18} & \textbf{85.50} \\
CLIP \cite{radford2021learning} & 89.09 & 90.48 & 90.86 & 77.93 & 81.29 & 82.98 \\
MAE \cite{he2022masked} & 87.26 & 88.75 & 90.30 & 76.51 & 81.84 & 82.39 \\
BEiT \cite{bao2021beit} & 85.64 & 88.74 & 89.71 & - & - & - \\
FaRL \cite{zheng2022general} & \underline{89.66} & \textbf{90.99} & \underline{91.39} & \textbf{78.77} & 82.57 & 83.75 \\ \midrule
Ours & \textbf{89.73} & \underline{90.89} & \textbf{91.74} & \underline{78.65} & \underline{83.15} & \underline{85.01} \\ \bottomrule
\end{tabular}}
\label{image_cmp}
\end{table}

We evaluate GPTFace in few-shot settings by fine-tuning its head with different proportions of training data, and the results are reported in Table~\ref{image_cmp}. Overall, our GPTFace outperforms state-of-the-art pre-training models in most scenarios. On CelebA dataset, GPTFace achieves higher accuracy than general pre-training transformer models such as ViT, CLIP, MAE, and BEiT. When the training ratio is 10\%, the face domain pre-training model, FaRL achieves 0.1\% higher accuracy than our GPTFace. On RAF-DB dataset, ViT achieves the best performance when training ratio is set as 50\% and 100\%, suggesting that pre-training on multi-class classification tasks (e.g., Imagenet22k) can benefit the fine-tuning of ViT for expression recognition. Despite being ranked as the second-best approach for all three ratios, GPTFace outperforms the face domain pre-training model, FaRL, on RAF-DB. Moreover, as a generative model, GPTFace can be applied to a variety of face editing tasks, while FaRL is only applicable to face analysis tasks.

\begin{figure*}[h]
\centerline{\includegraphics[width=\textwidth]{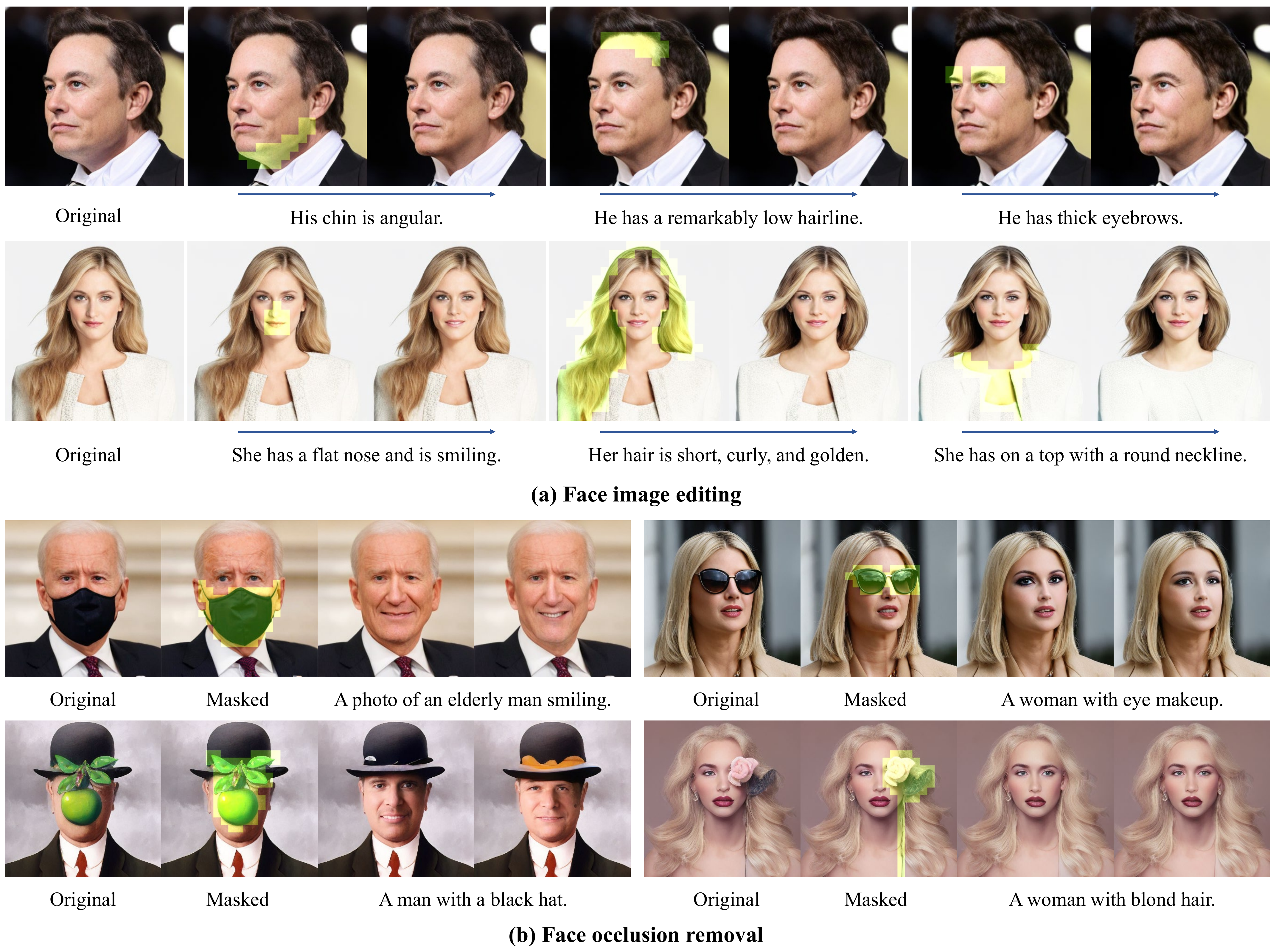}}
\caption{Results of text-guided face editing. The masked region is highlighted in green, and the model uses the provided text to generate predictions for the masked areas.}
\label{exp_1}
\end{figure*}

\subsection{Text-guided Face Editing}

Different from previous methods, we achieve face editing with the transformer pre-training model for the first time. As shown in Figure~\ref{exp_1}, we mask a portion of the image and ask the model to predict the masked region based on the text description. Figure~\ref{exp_1}(a) shows the results of continuous face attribute editing using the proposed approach, which enables unaligned face image editing without pre-alignment, typically required in traditional GAN-based editing methods. Moreover, our approach also allows the direct handling of the original image without the inversion process \cite{leng2021force,zhu2020domain,abdal2020image2stylegan++,abdal2019image2stylegan}.

Our approach leverages large-scale general face data and can also be used for editing beyond facial attributes, including clothing style modification, making it highly versatile. Figure~\ref{exp_1}(a) demonstrates the effectiveness of our approach for face occlusion removal, which can restore a natural-looking appearance and can even be applied to paintings.

We compare our model to both StyleGAN-based and Diffusion-based facial editing methods, including StyleCLIP \cite{patashnik2021styleclip}, Stable Diffusion \cite{rombach2021highresolution}, and StyleCLIP with FaRL \cite{zheng2022general} guided mapper. Our experiments are conducted using aligned and unaligned face images, all scaled to $256\times 256$ resolution. Regarding the experiment settings, we adopt the latent optimization approach of StyleCLIP, and employ inpainting to manipulate masked regions for Stable Diffusion. Notably, we used the same mask as that used in the stable diffusion in our experiments.

\begin{figure}[h]
\centerline{\includegraphics[width=\linewidth]{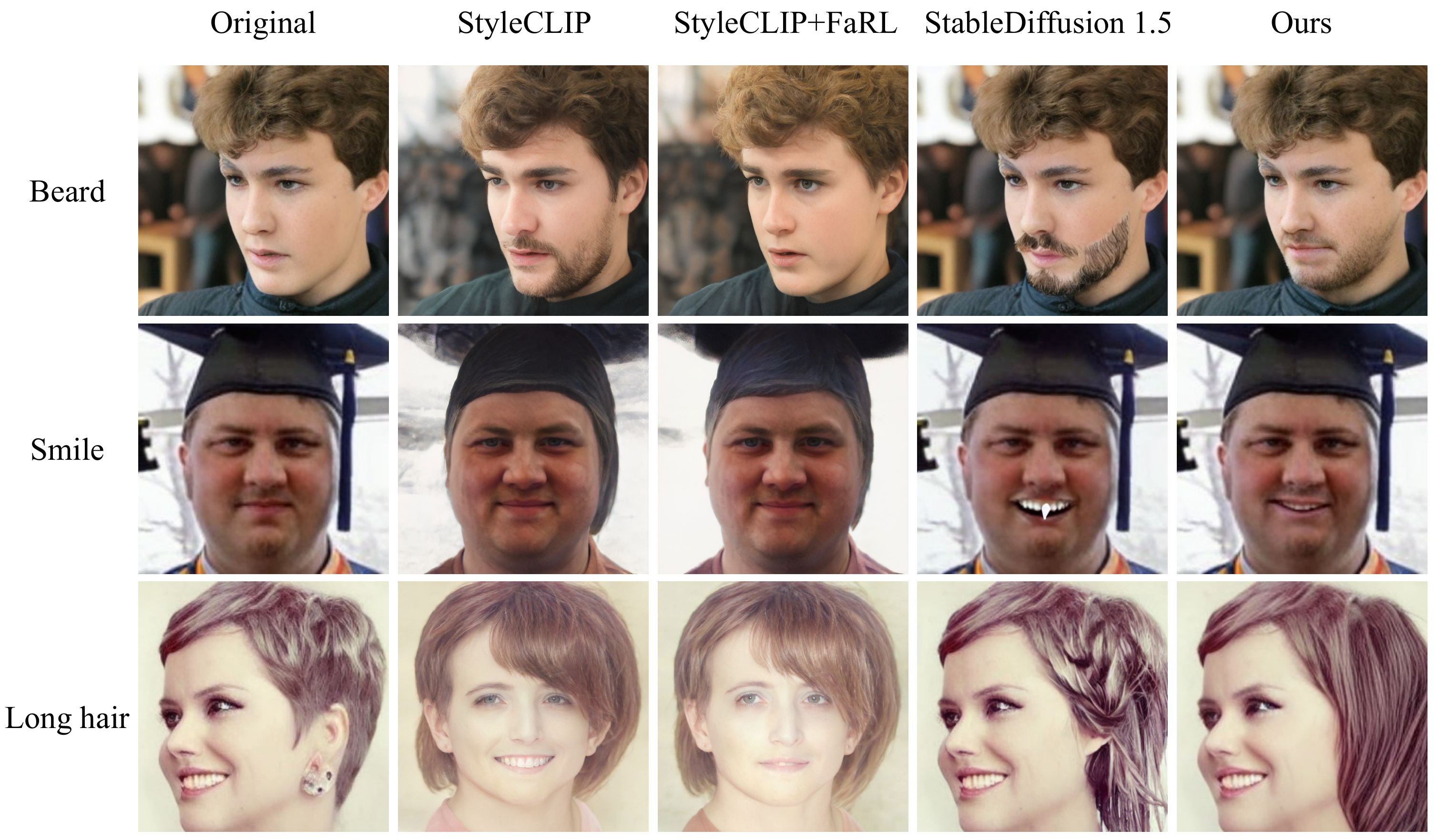}}
\caption{A comparison with StyleCLIP \cite{patashnik2021styleclip}, FaRL \cite{zheng2022general}, and Stable Diffusion \cite{rombach2021highresolution}.}
\label{exp_2}
\end{figure}

As shown in Figure~\ref{exp_2}, our method demonstrates the ability to perform text-guided editing with minimal disruption to the surrounding area of the edit. In contrast, StyleGAN-based methods require aligned faces as input and may suffer from information loss during inversion. Latent level editing can also lead to attribute entanglement, as demonstrated by the results of StyleCLIP and FaRL, which exhibit changes in the face identity in the first row, failure to reconstruct the hat in the second row, and unsuccessful inversion in the third row. Although Stable Diffusion has advantages in spatial decoupling, the generated results may contain artifacts, such as teeth in the second row. Additionally, FaRL is a pre-trained model for face analysis tasks that depends on methods like StyleCLIP for image editing. However, as evidenced by the FaRL for beard editing in the first row, it is not always effective. In contrast, our method leverages a large-scale pre-training dataset and can support multiple editing approaches and face analysis tasks simultaneously, making it more versatile and effective.

\subsection{Image-guided Text Generation}

Our model goes beyond face editing and can also achieve image-guided text generation. In this section, we evaluate our model's performance on the image tagging task, which involves generating a list of textual tags (keywords) from an input image.

\begin{figure}[t]
\centerline{\includegraphics[width=\linewidth]{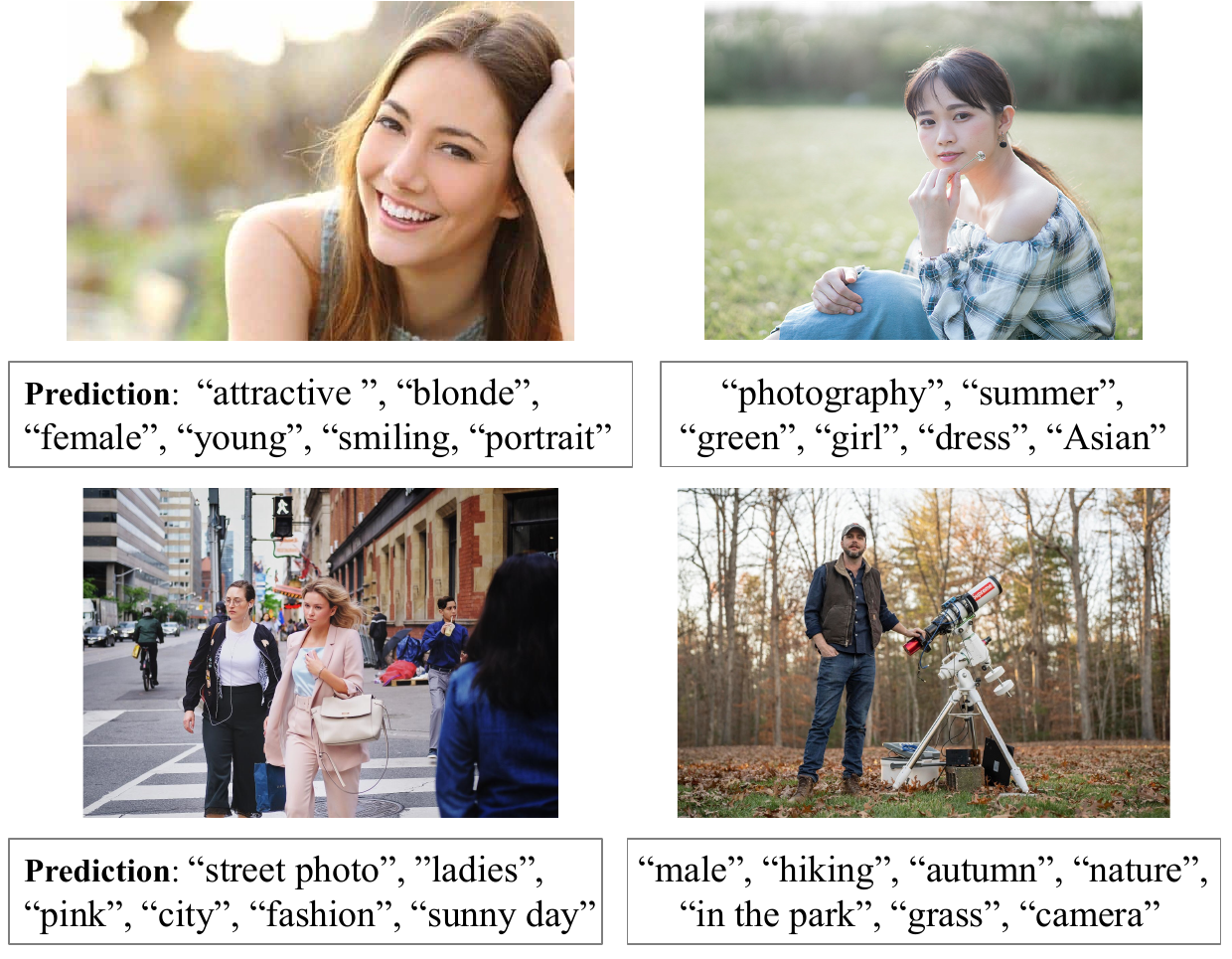}}
\caption{Results of face image tagging.}
\label{taggings}
\end{figure}

To obtain the predicted tags, we use a template of the form \textit{``Tags: \{MASK\}, \{MASK\}, ...''} and prompt the model to predict the tags based on the given image. Table~\ref{taggings} presents examples of our model's prediction, demonstrating its ability to accurately predict tags for each face image. To compare our approach with existing pretraining-based image tagging methods, we conduct a user study using 10 random face images from Google Search as test data.

Following the evaluation settings used in previous studies \cite{huo2021wenlan}, we generate 30 results for each test image and collect feedback from human evaluators via a Google Form. Evaluators are asked to rate the results using a three-point scale: 0, 1, and 2. To recruit volunteers, we reach out to individuals through university email lists and social media platforms.

To evaluate the human retrieval quality, we adopt the NDCG and mAP metrics, which are commonly used to assess the accuracy of retrieval. As shown in Table~\ref{user_study}, our method outperforms BriVL \cite{huo2021wenlan} and CLIP \cite{radford2021learning}. It is worth noting that the comparison methods are retrieval-based and select tags from candidates, while our method directly generates textual tags, which is more practical and better suited for real-world scenarios.

\begin{table}[h]
\centering
\small
\caption{User study results for the image-guided text generation task.}
\begin{tabular}{l|llll}
\toprule
 & \multicolumn{1}{c}{NDCG@5} & \multicolumn{1}{c}{NDCG@10} & \multicolumn{1}{c}{NDCG@20} & \multicolumn{1}{c}{mAP} \\ \midrule
CLIP  & 32.9 & 38.8 & 53.0 & 30.3 \\
BriVL  & 37.5 & 42.8 & 55.5 & 37.6 \\ \midrule
Ours & 43.7 & 46.7 & 68.8 & 40.2 \\ \bottomrule
\end{tabular}
\label{user_study}
\end{table}

\section{Further Analysis}

\subsection{Effectiveness of VQGAN}

The ability of VQGAN to faithfully reconstruct the original image is a crucial indicator when utilizing VQGAN as a discrete image representation for image editing. For example, for face generation, VQGAN is simply required to produce realistic images, but for face editing, the output images must retain the characteristics of the input exemplar. This places higher requirements on the reconstruction ability of the VQGAN models.

\begin{figure}[h]
\centerline{\includegraphics[width=\linewidth]{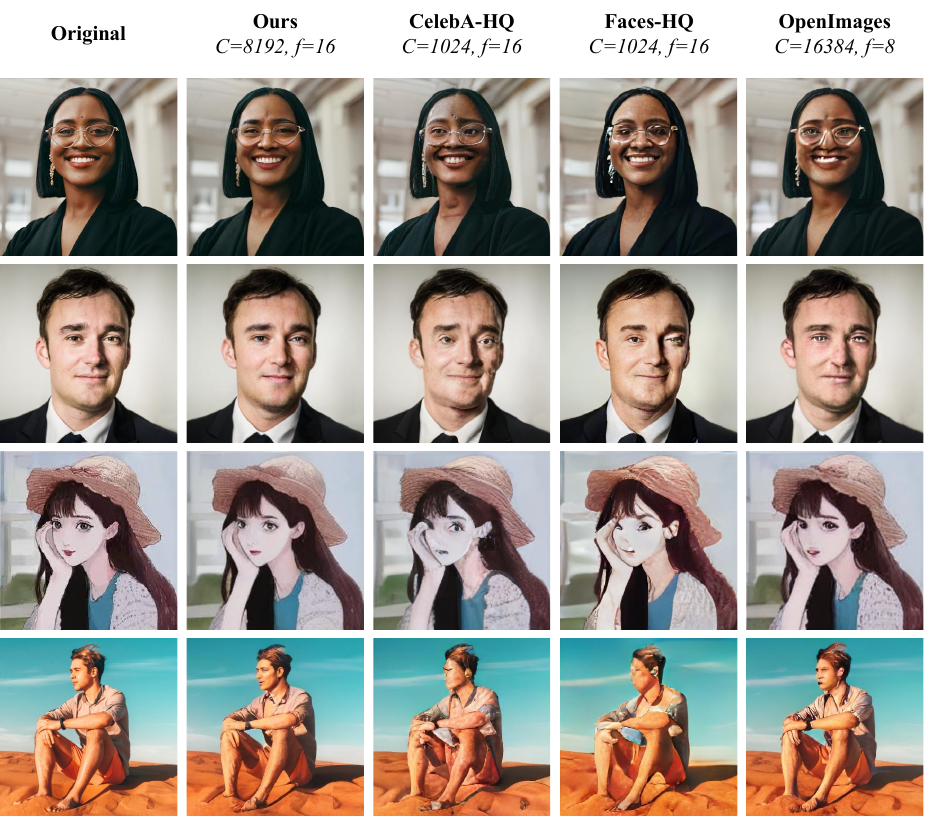}}
\caption{Comparison of reconstruction results between our VQGAN with other publicly available VQGANs trained on different datasets.}
\label{exp_4}
\end{figure}

In this study, we train a VQGAN model from scratch on the LAION-FACE dataset, as publicly available VQGAN models are found to struggle with face editing tasks. Our VQGAN model, as shown in Figure~\ref{exp_4}, outperforms other versions in preserving the original attributes of the face image. For instance, when comparing the results of the first two rows, other models tend to alter features such as the shape of the woman's glasses and the eyes of the men, while our model retains these characteristics. Moreover, our model produces clearer images for face images with diverse styles, such as paintings and unaligned portraits. In contrast, other models tend to produce blurry images.

\subsection{Unconditional Face Inpainting}

\begin{figure}[t]
\centerline{\includegraphics[width=\linewidth]{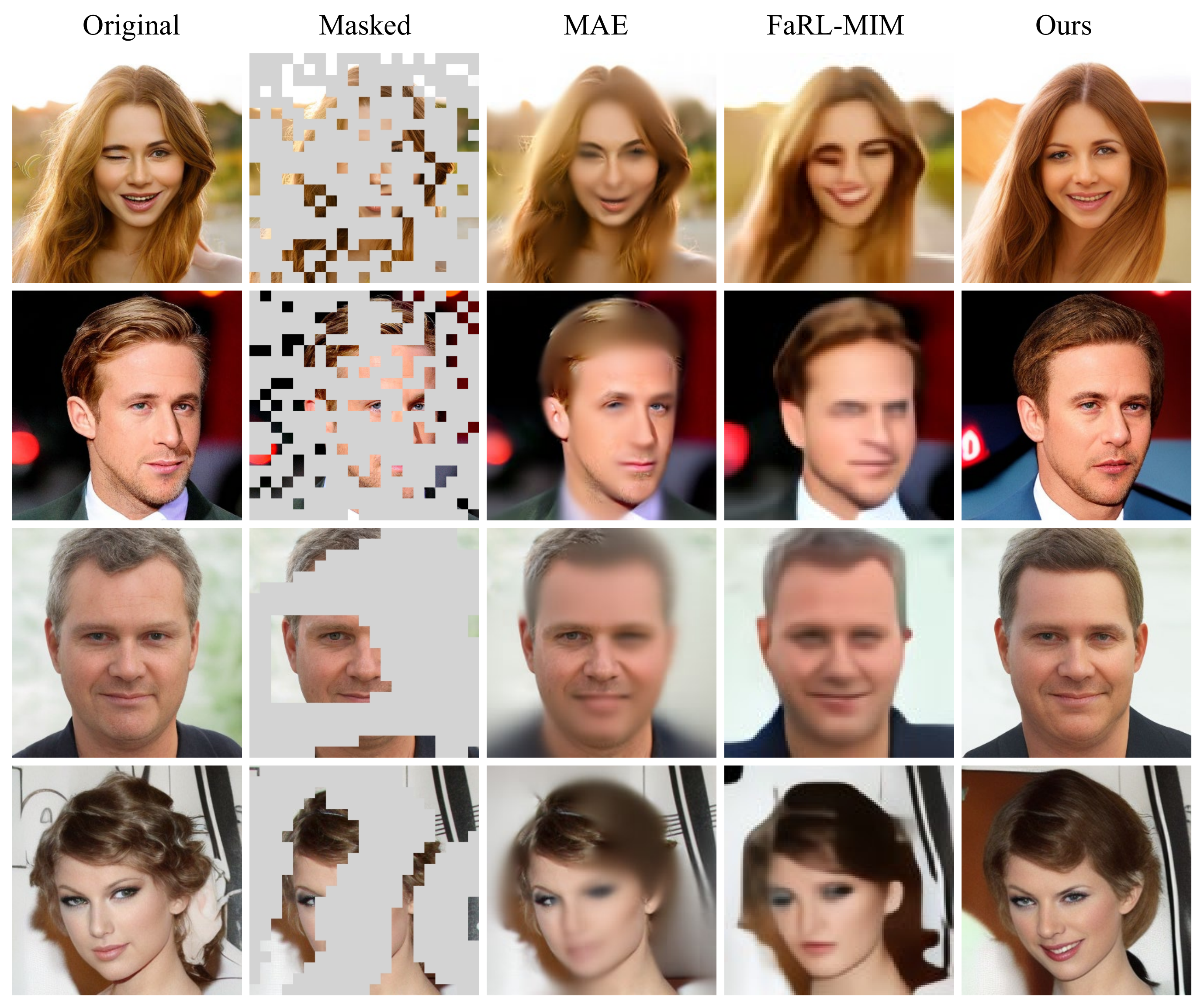}}
\caption{Comparison with MAE and FaRL for face inpainting.}
\label{mask_exp}
\end{figure}

Our model for image inpainting reconstructs missing parts of an image based on the underlying face distribution. In Figure~\ref{mask_exp}, we present the results of our experiments using two masking approaches: random masking (first two rows) and span masking (last two rows). To compare the performance of our approach, we evaluate the masked autoencoders (MAE) \cite{he2022masked} and FaRL with masked image modeling (FaRL-MIM) \cite{zheng2022general} using a 75\% random masking ratio.

Our experiments demonstrate that while MAE can restore the shape and color of face images, it often results in blurry facial features. In addition, the images recovered by FaRL-MIM appear to be blurry and lack the necessary texture information. These findings are observed in both the random and span masking experiments. In contrast, our method not only restores the shape and color of facial features but also preserves precise texture details for both masking approaches.

\begin{figure}[h!]
\centerline{\includegraphics[width=\columnwidth]{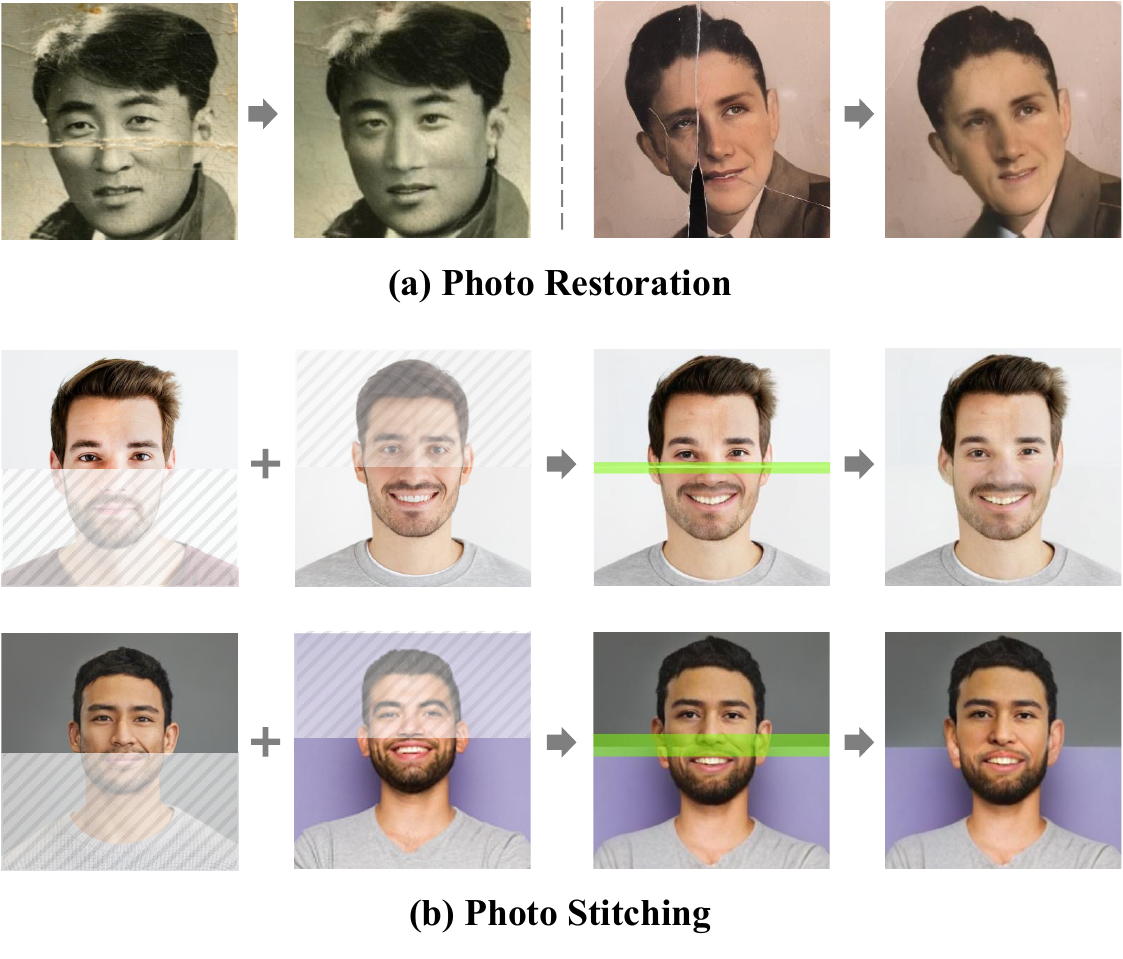}}
\caption{Examples of photo restoration and stitching.}
\label{exp_3}
\end{figure}

The model's ability to accurately preserve texture details has significant potential for various real-world applications. To demonstrate this, we conduct additional experiments on different unconditional generation tasks. For the photo restoration, we mask scratches or cracks in the image and task the model with predicting the masked positions. Figure~\ref{exp_3}(a) showcases how our model successfully repairs corrupted photos. We further test our model by using it to stitch two face photos, masking the stitched edges, and tasking the model with restoring them. The results, depicted in Figure~\ref{exp_3}(b), demonstrate the model's effectiveness in repairing the traces of stitching.

\section{Conclusion}

This paper introduces GPTFace, a generative pre-training model that learns facial knowledge from a large-scale, weakly correlated dataset of texts and images. Additionally, we propose a gradient-based controllable generation approach for text/image-guided image/text generation. GPTFace is versatile and can be applied to various face-related tasks, such as face editing, face image tagging, and facial analysis. Compared to state-of-the-art pre-training models, our model achieves comparable performance and is applicable to more scenarios. We will make our code and pre-trained models publicly available.

\textbf{Limitations and future work.} One limitation of our approach is that the masked area needs to be provided in advance. However, the impressive zero-shot segmentation performance of the SAM model \cite{kirillov2023segment} may offer a potential solution to this challenge. In future work, we plan to investigate more sophisticated text-to-image matching techniques to automatically determine the areas that require editing in images. Furthermore, we aim to incorporate transformer variants with linear time complexity to accelerate training for longer sequence lengths and high-resolution face images.

\bibliography{custom,sample-base}

\end{document}